\documentclass[sigconf]{acmart}

\usepackage{amsmath,amsfonts}
\usepackage{algorithmic}
\usepackage{graphicx}
\usepackage{textcomp}
\usepackage{xcolor}
\usepackage{multirow}
\usepackage{algorithm}
\usepackage{algorithmic}
\usepackage{hyperref}
\usepackage{url}

\usepackage{subfigure}
\newcommand{\jiaxing}[1]{{\color{black}#1}}

\AtBeginDocument{%
  }

\copyrightyear{2024}
\acmYear{2024}
\setcopyright{rightsretained}
\acmConference[CIKM '24]{Proceedings of the 33rd ACM International Conference on Information and Knowledge Management}{October 21--25, 2024}{Boise, ID, USA}
\acmBooktitle{Proceedings of the 33rd ACM International Conference on Information and Knowledge Management (CIKM '24), October 21--25, 2024, Boise, ID, USA}
\acmDOI{10.1145/3627673.3679560}
\acmISBN{979-8-4007-0436-9/24/10}

\makeatletter
\gdef\@copyrightpermission{
  \begin{minipage}{0.3\columnwidth}
\href{https://creativecommons.org/licenses/by/4.0/}{\includegraphics[width=0.90\textwidth]{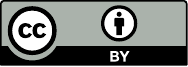}}
  \end{minipage}\hfill
  \begin{minipage}{0.7\columnwidth}
   \href{https://creativecommons.org/licenses/by/4.0/}{This work is licensed under a Creative Commons Attribution International 4.0 License.}
  \end{minipage}
}
\makeatother

\begin{document}

\title{Contrasformer: A Brain Network Contrastive Transformer for Neurodegenerative Condition Identification}

\author{Jiaxing Xu}
\authornote{Corresponding Author}
\affiliation{%
  \institution{Nanyang Technological University}
  \country{Singapore}
}
\email{jiaxing003@e.ntu.edu.sg}

\author{Kai He}
\affiliation{%
  \institution{National University of Singapore}
  \country{Singapore}}
\email{kai_he@nus.edu.sg}

\author{Mengcheng Lan}
\affiliation{
  \institution{S-Lab}
  \city{Nanyang Technological University}
  \country{Singapore}
}
\email{lanm0002@e.ntu.edu.sg}

\author{Qingtian Bian}
\affiliation{
  \institution{Nanyang Technological University}
  \country{Singapore}
}
\email{bian0027@ntu.edu.sg}

\author{Wei Li}
\affiliation{
  \institution{Nanyang Technological University}
  \country{Singapore}
}
\email{wei014@e.ntu.edu.sg}

\author{Tieying Li}
\affiliation{
  \institution{Northeastern University}
  \city{Shenyang}
  \country{China}
}
\email{tieying@stumail.neu.edu.cn}

\author{Yiping Ke}
\affiliation{
  \institution{Nanyang Technological University}
  \country{Singapore}
}
\email{ypke@ntu.edu.sg}

\author{Miao Qiao}
\affiliation{
  \institution{The University of Auckland}
  \city{Auckland}
  \country{New Zealand}
}
\email{miao.qiao@auckland.ac.nz}

\renewcommand{\shortauthors}{Jiaxing Xu et al.}

\begin{abstract}
Understanding neurological disorder is a fundamental problem in neuroscience, which often requires the analysis of brain networks derived from functional magnetic resonance imaging (fMRI) data.
Despite the prevalence of Graph Neural Networks (GNNs) and Graph Transformers in various domains, applying them to brain networks faces challenges. 
Specifically, the datasets are severely impacted by the noises caused by distribution shifts across sub-populations and the neglect of node identities, both obstruct the identification of disease-specific patterns. To tackle these challenges, we propose \textit{Contrasformer}, a novel contrastive brain network Transformer. \jiaxing{It generates a prior-knowledge-enhanced contrast graph to address the distribution shifts across sub-populations by a two-stream attention mechanism.} A cross attention with identity embedding highlights the identity of nodes, and three auxiliary losses ensure group consistency. Evaluated on 4 functional brain network datasets over 4 different diseases, Contrasformer outperforms the state-of-the-art methods for brain networks by achieving up to 10.8\% improvement in accuracy, which demonstrates its efficacy in neurological disorder identification. Case studies illustrate its interpretability, especially in the context of neuroscience. This paper provides a solution for analyzing brain networks, offering valuable insights into neurological disorders. 
Our code is available at \url{https://github.com/AngusMonroe/Contrasformer}.
\end{abstract}


\maketitle

\section{Introduction}
\label{sec:introduction}

In the field of neuroscience, a central objective is to uncover distinctive patterns associated with neurological disorders (e.g., Alzheimer’s, Parkinson’s, and Autism) by exploring the brain networks of both healthy individuals and patients with neurological disorders \cite{poldrack2009decoding}. Among the various techniques for neuroimaging, resting-state functional magnetic resonance imaging (fMRI) is widely used to profile the connectivities among brain regions \cite{worsley2002general}, leading to brain networks where each node is a specific brain region,  called a region of interest (ROI), and each edge denotes a pairwise correlation between the blood-oxygen-level-dependent (BOLD) signals of two ROIs \cite{xu2023data}. The edge reveals the connectivity between brain regions, showcasing which areas tend to be activated together or exhibit correlated activities. Brain networks model neurological systems as graphs and one could apply graph-based techniques to gain insights into their roles and interactions \cite{kawahara2017brainnetcnn,lanciano2020explainable,wang2023effective}. Such brain networks could help to decipher distinctive patterns associated with neurological disorders, which can contribute to early diagnosis and effective intervention strategies. It's worth noting that the development of graph analytics for brain networks is still in its early stages.

While graph neural networks (GNNs)\cite{kipf2016semi,hamilton2017inductive,velivckovic2017graph} and graph Transformers \cite{vaswani2017attention,ying2021transformers,rampavsek2022recipe} have recently been adopted in a wide range of graph-related tasks, applying them to brain networks faces two challenges below in capturing the disease-specific pattern.

\begin{figure}[h]
  \centering
\subfigure[\begin{scriptsize}\textbf{Site distribution for original feature.}\end{scriptsize}]{
    \includegraphics[scale=0.22]{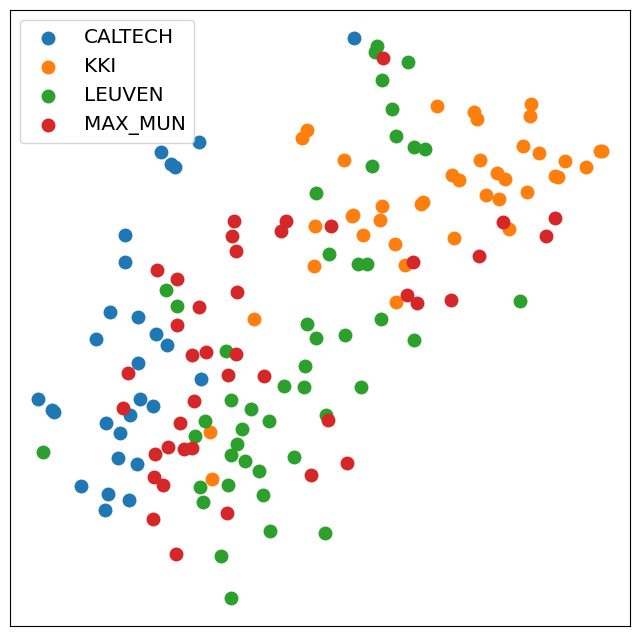}
    }
    \subfigure[\begin{scriptsize}\textbf{Scanning duriation distribution for original feature.}\end{scriptsize}]{
    \includegraphics[scale=0.22]{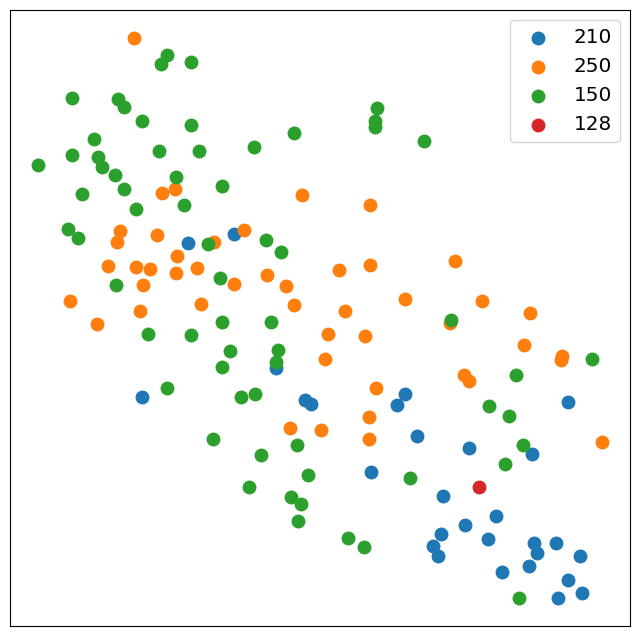}
    }
    \subfigure[\begin{scriptsize}\textbf{Site distribution for Contrasformer representation.}\end{scriptsize}]{
    \includegraphics[scale=0.22]{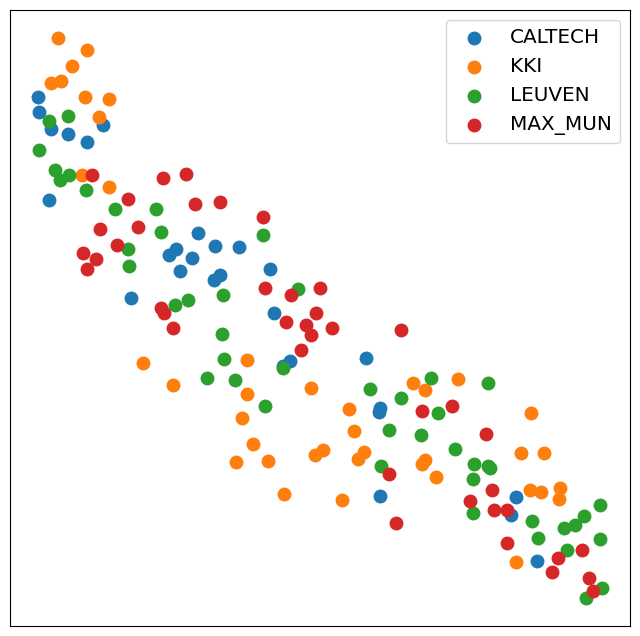}
    } 
    \subfigure[\begin{scriptsize}\textbf{Scanning duriation distribution for Contrasformer representation.}\end{scriptsize}]{
    \includegraphics[scale=0.22]{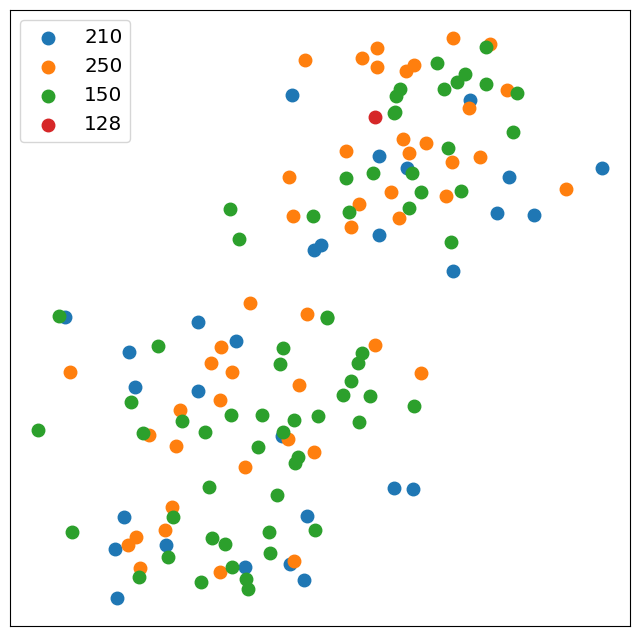}
    } 
  \caption{The distribution of the original feature and Contrasformer representation for subjects from multiple sites and scanning duriation in ABIDE dataset. Each point in the figure represents a subject and different colors denote the sites these subjects are acquired from or their lengths of the BOLD signals. The representation of each subject is obtained by mean pooling and visualized by t-SNE \cite{van2008visualizing}. Compared with (c) and (d), (a) and (b) exhibit obvious distribution shifts.}
  \label{fig:feat_dist}
\end{figure}

\textbf{Sub-population-specific (SPS) Noise.} Analyzing neurological disorders aims to capture disease-specific patterns that are invariant across all populations. However, certain features are common to a sub-population (not generalizable to the entire population) and are unrelated to the disease, constituting sub-population-specific (SPS) noise. For instance, brain network datasets like Autism Brain Imaging Data Exchange (ABIDE) \cite{craddock2013neuro} and Alzheimer’s Disease Neuroimaging Initiative (ADNI) \cite{dadi2019benchmarking} are collected from multiple sites. Subjects from different sites may exhibit site differences (scanner variability, different inclusion/exclusion criteria)\cite{chan2022semi}. Such noise could lead the model to focus on the site-specific pattern instead of learning population-invariant information. Besides, the varying scanning duration could result in different periods of region activation recorded for subjects. Furthermore, label inconsistencies may arise due to differences in diagnostic criteria used by doctors labeling these subjects.
Fig. \ref{fig:feat_dist}(a) provides an example of the site distribution of subjects in the ABIDE dataset. Each point in the figure represents a subject, and different colors denote the sites from which these subjects are acquired. From this example, it can be observed that subjects from the same site tend to have similar features, as the site-specific noise dominates the similarity. A similar observation is found in the distribution of sub-populations with different scanning durations. As illustrated in Fig. \ref{fig:feat_dist}(b), each point represents a subject, and different colors denote the lengths of the BOLD signal for these subjects. Such a conspicuous distribution shift across sub-populations could easily mislead the model to overfit the SPS noise, thereby limiting its performance.

\textbf{Node-identity awareness.} The construction of brain networks requires a specific parcellation method to split the whole brain into several ROIs. The same parcellation method is applied to all subjects, and thus the ROI definition is identical across all brain networks. Such a property does not generally exist in other graph-structured data, necessitating our specialized tailoring for brain networks. 
Existing general-purposed GNNs are designed to learn the structural pattern of graphs without considering their node identities \cite{velivckovic2017graph,xu2018powerful,xu2024union}.

While some recent GNNs have introduced specialized designs for brain networks and addressed issues related to node identity \cite{li2020pooling,li2021braingnn,zhang2022classification}, many of them overlook addressing SPS noise and capturing group-specific patterns. Neglecting these aspects can lead to models that overfit outliers and hinder interpretability \cite{kan2022fbnetgen}. 

In this paper, we propose \textit{Contrasformer}, a novel contrastive brain network Transformer, that harnesses the distinctive properties of brain networks to fully leverage the capabilities of Transformer-based models for brain network analysis. Specifically, we employ a two-stream attention mechanism to generate a contrast graph that encodes group-specific information. Such two-stream attention mechanism can address the SPS noise by reweighting the original feature across ROIs and subjects. The distribution shift will also be corrected as shown in Fig. \ref{fig:feat_dist}(c) and (d). Meanwhile, an identity embedding is introduced to incorporate the ROI identities with the input brain network, which highlights the node identity awareness. Moreover, we propose a cross decoder to integrate the contrast graph with the encoded brain network using a cross-attention mechanism. We further take advantage of node identity awareness by introducing a contrastive loss to constrain that identical ROIs across subjects have similar representations. Additionally, a cluster loss is introduced to guarantee group consistency in graph representations and take the graph-level relationship into account. Our contributions can be summarized as follows:

\begin{itemize}
\item We introduce a contrast graph encoder to adaptively assign weights across ROIs and subjects by a two-stream attention mechanism. The generated contrast graph includes the most discriminative ROIs concerning different groups to address sub-population distribution shift and remit overfitting.
\item By incorporating the contrast graph with brain network representation learning, we propose a contrastive brain network Transformer named \textit{Contrasformer}. It utilizes an identity encoding and three auxiliary losses to make the model aware of node identity and graph-level relationship.
\item We evaluate Contrasformer on 4 fMRI brain network datasets for different neurological disorders. Our results demonstrate the superiority of Contrasformer over 13 state-of-the-art methods on neurological disorder identification.
\item We present a case study that underscores the intriguing, straightforward, and highly interpretable patterns extracted by our approach, aligning with domain knowledge found in neuroscience literature.
\end{itemize}

\section{Related Work}
\label{sec:related_work}

\subsection{Graph Neural Networks (GNNs)}

GNNs have gained prominence as a popular approach in processing and analyzing graph structure data, including social networks \cite{kipf2016semi}, molecular data \cite{gilmer2017neural}, knowledge graphs \cite{li2023imf,wu2023megacare} and user-item graphs \cite{bian2023cpmr}. Most GNNs operate on a message-passing scheme \cite{kipf2016semi,hamilton2017inductive,velivckovic2017graph}, wherein they iteratively aggregate information from neighboring nodes and update the representation of each node \cite{xu2018powerful}. 
In recent years, there has been a surge of interest in employing GNNs for the analysis of brain networks. Ktena et al. \cite{ktena2017distance} utilized graph convolutional networks to learn similarities between pairs of brain networks (subjects). BrainNetCNN \cite{kawahara2017brainnetcnn} introduced edge-to-edge, edge-to-node, and node-to-graph convolutional filters to harness the topological information within brain networks. LiNet \cite{li2019graph} presented a two-stage pipeline that uses GNNs to identify brain biomarkers associated with Autism Spectrum Disorder (ASD) from fMRI data. BrainGNN \cite{li2021braingnn} proposed an ROI-selection pooling method to emphasize salient brain regions for each individual. However, these approaches often overlook the unique characteristics of functional brain networks, such as node-identity awareness \cite{lanciano2020explainable}.

To better utilize node-identity awareness of brain networks, PRGNN \cite{li2020pooling} introduced a graph pooling method with group-level regularization, in the mean time ensuring group-level consistency. GroupINN \cite{yan2019groupinn} jointly learnt node grouping and graph feature extraction, though without considering node alignment. Lanciano et al. \cite{lanciano2020explainable} focused on extracting a dense contrast subgraph to filter relevant information for predictions. Nevertheless, their feature extraction and subject classification treated all brain regions and individuals uniformly, potentially making them susceptible to noisy data. Zhang et al. \cite{zhang2022classification} incorporated both local ROI-GNN and global subject-GNN guided by non-imaging data, but the local ROI-GNN did not account for the node alignment in brain networks. ContrastPool \cite{10508252}, introduced a dual attention mechanism to extract discriminative features across ROIs for subjects within the same group. However, it still neglects the graph-level relationship between different groups, which could suffer from the SPS noise.

In addition, a general issue for GNNs based on the message-passing scheme is called over-smoothing \cite{keriven2022not}, which refers to the loss of discriminative information as the network iteratively aggregates information from a large number of neighbors. In our context of brain networks, correlation-based graphs naturally possess high density that could easily suffer from over-smoothing. Therefore we discard message-passing architecture in our work.

\subsection{Graph Transformer}

An alternative method for graph representation learning involves Transformer-based models \cite{vaswani2017attention}, which adapt the attention mechanism to consider global information for each node and incorporate positional encoding to capture graph topological information.  Graph Transformers have garnered significant attention due to their impressive performance in graph representation learning. Dwivedi et al. \cite{dwivedi2020generalization} introduced edge information into the attention mechanism and used eigenvectors as positional embeddings. SAN  \cite{kreuzer2021rethinking} implemented an invariant aggregation of Laplacian's eigenvectors for positional embedding and introduced conditional attention for real and virtual edges within a graph. Graphormer \cite{ying2021transformers} enhanced the attention mechanism with centrality-based positional embedding and introduced pair-wise graph distances to define relative positional encodings. More recently, GPS \cite{rampavsek2022recipe} proposed a hybrid architecture that combines GNN and Transformer components, achieving state-of-the-art results on various datasets by introducing different types of global/local/relative positional/structural embeddings.

Nonetheless, applying these Transformer-based models to brain networks presents challenges \cite{kan2022brain}, primarily due to the correlation-based edges that hinder the use of designs like centrality \cite{dwivedi2021graph}, spatial \cite{ying2021transformers}, and edge encoding \cite{rampavsek2022recipe}. Consequently, a series of brain network Transformer methods have emerged. One such method \cite{kan2022brain} applied Transformers to learn pairwise connection strengths among brain regions across individuals.  THC \cite{dai2023transformer} introduced an interpretable Transformer-based model for joint hierarchical cluster identification and brain network classification. DART \cite{kan2023dynamic} utilized segmenting BOLD signals to generate dynamic brain networks and then incorporated them with static networks for representation learning. Nevertheless, these approaches often neglect the incorporation of group-level information from subjects into their methodology design. Thus, it would be hard to address the issue of SPS noise. \jiaxing{The lack of leveraging node-identity awareness can also limit the power of Transformer.}

\begin{figure*}[h]
\begin{center}
\includegraphics[width=.75\linewidth]{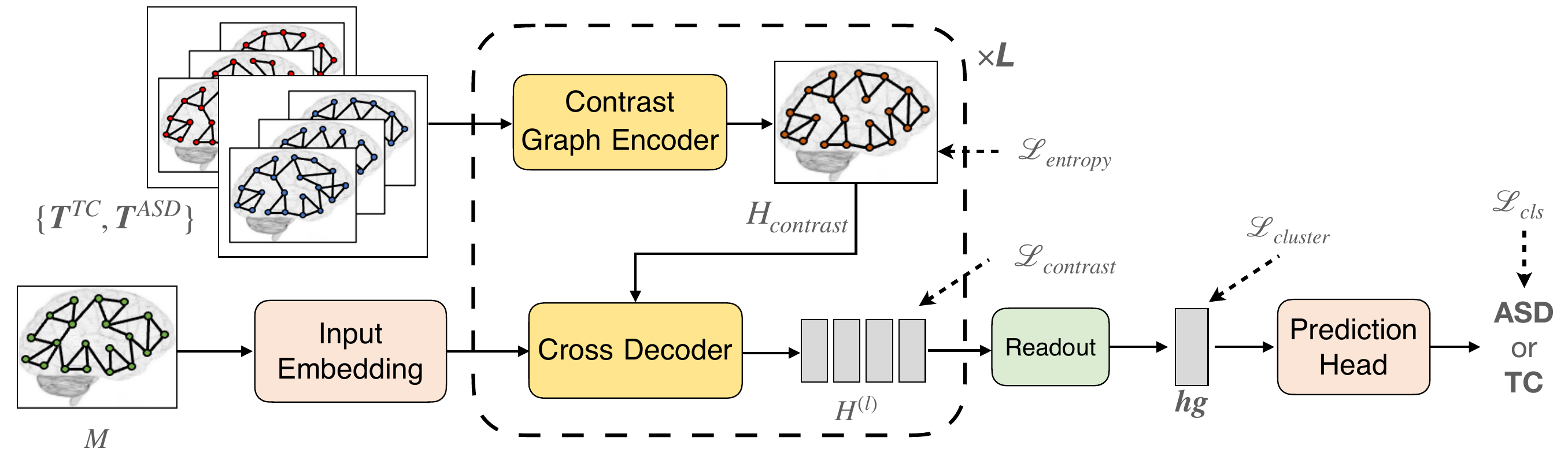}
\end{center}
\caption{The framework of Contrasformer for neurological disorder identification, using Autism as an example.}
\label{fig:framework}
\end{figure*}

\section{Preliminaries}
\label{sec:preliminaries}

\subsection{Brain Network Construction}

In this work, we use the datasets preprocessed and released by Xu et al \cite{xu2023data}. We apply Schaefer atlas \cite{schaefer2018local} to parcellate all subjects, which divides the brain into 100 functional ROIs.  
For each subject, a brain network is represented by a connectivity matrix $\boldsymbol{M}\in \mathbb{R}^{m \times m}$, where $m$ denotes the number of ROIs. The nodes in $\boldsymbol{M}$ are ROIs and the edges are the Pearson's correlation between the region-averaged BOLD signals from pairs of ROIs. Essentially, $\boldsymbol{M}$ captures functional relationships between ROIs.
For the detailed preprocessing and brain network construction pipeline, please refer to Xu et al~\cite{xu2023data}.

\begin{figure*}[h]
\begin{center}
\includegraphics[width=.8\linewidth]{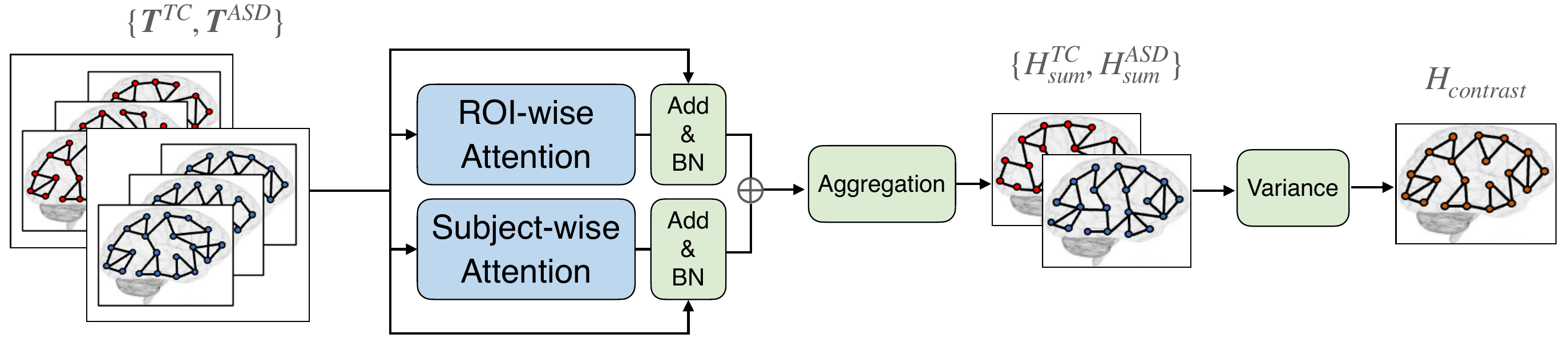}
\end{center}
\caption{The architecture of contrast graph encoder. Each group of brain networks is fed into the two-stream attention to obtain a summary graph. The contrast graph is generated by contrasting the summary graphs of different groups.}
\label{fig:encoder}
\end{figure*}

\subsection{Problem Definition}

Neurological disorder identification for brain networks aims to predict the distinct class of each subject. Given a dataset of labeled brain networks $\mathcal{D} = (\mathcal{M}, \mathcal{Y}) = \{(\boldsymbol{M}, y_M)\}$, where $y_M$ is the class label of a brain network $\boldsymbol{M} \in \mathcal{M}$, the problem of neurological disorder identification is to learn a predictive function $f$: $\mathcal{M} \rightarrow \mathcal{Y}$, which maps input brain networks to the groups they belong to, expecting that $f$ also works well on unseen brain networks. We use ABIDE dataset, which contains groups of typical controls (TC) and individuals diagnosed with ASD, as an example to explain our methodology in the following.
Notation-wise, we use calligraphic letters to denote sets (e.g., $\mathcal{M}$), bold capital letters to denote matrices (e.g., $\boldsymbol{M}$), and strings with bold lowercase letters to represent vectors (e.g., $\boldsymbol{hg}$).  Subscripts and superscripts are used to distinguish between different variables or parameters, and lowercase letters denote scalars. We use $\boldsymbol{M}(i,:)$ and $\boldsymbol{M}(:,j)$ to denote the $i$-th row and $j$-th column of a matrix $\boldsymbol{M}$, respectively. This notation also extends to a 3d matrix. 

\section{Methodology}
\label{sec:method}

In this section, we provide a detailed exposition of the design of our proposed Contrasformer, depicted in Fig. \ref{fig:framework}. Contrasformer adopts an encoder-decoder architecture, featuring three key components: (1) A contrast graph encoder is introduced (in Section \ref{subsec:contrast_graph}) to extract the most discriminative task-related features from the training set. To alleviate the SPS noise, a two-stream attention mechanism is employed to generate a contrast graph that captures the invariant information across all sub-population. The learnt contrast graph is then utilized in both training and test stages to be incorporated with the brain network representation learning.
(2) A cross decoder is introduced (in Section \ref{subsec:contrasformer}) to combine the input brain network with node identity, and subsequently fuse the contrast graph with the identity-embedded brain network by a cross-attention to update node representations.
(3) A classification loss $\mathcal{L}_{cls}$ and three auxiliary losses $\mathcal{L}_{entropy}$, $\mathcal{L}_{cluster}$, $\mathcal{L}_{contrast}$ are incorporated (in Section \ref{subsec:loss}) to guide the end-to-end training. These losses emphasize the node identity of ROIs and consider group-level relationships.

\subsection{Contrast Graph Encoder with Two-stream Attention}
\label{subsec:contrast_graph}

To generate a contrast graph with group-specific information, we introduce a two-stream contrast graph encoder. In neuroscience, normally there are only some ROIs in the brain that can reflect the lesion of a neurological disorder. So the aim of this encoder is to extract the most discriminative ROIs for each group while adaptly learning the contribution of each subject. \jiaxing{Such a two-stream attention design is able to capture the population-invariant information embedded in subjects and ROIs, which alleviates the impact of SPS noise in the downstream task.}

\begin{figure}[h]
\begin{center}
\includegraphics[width=\linewidth]{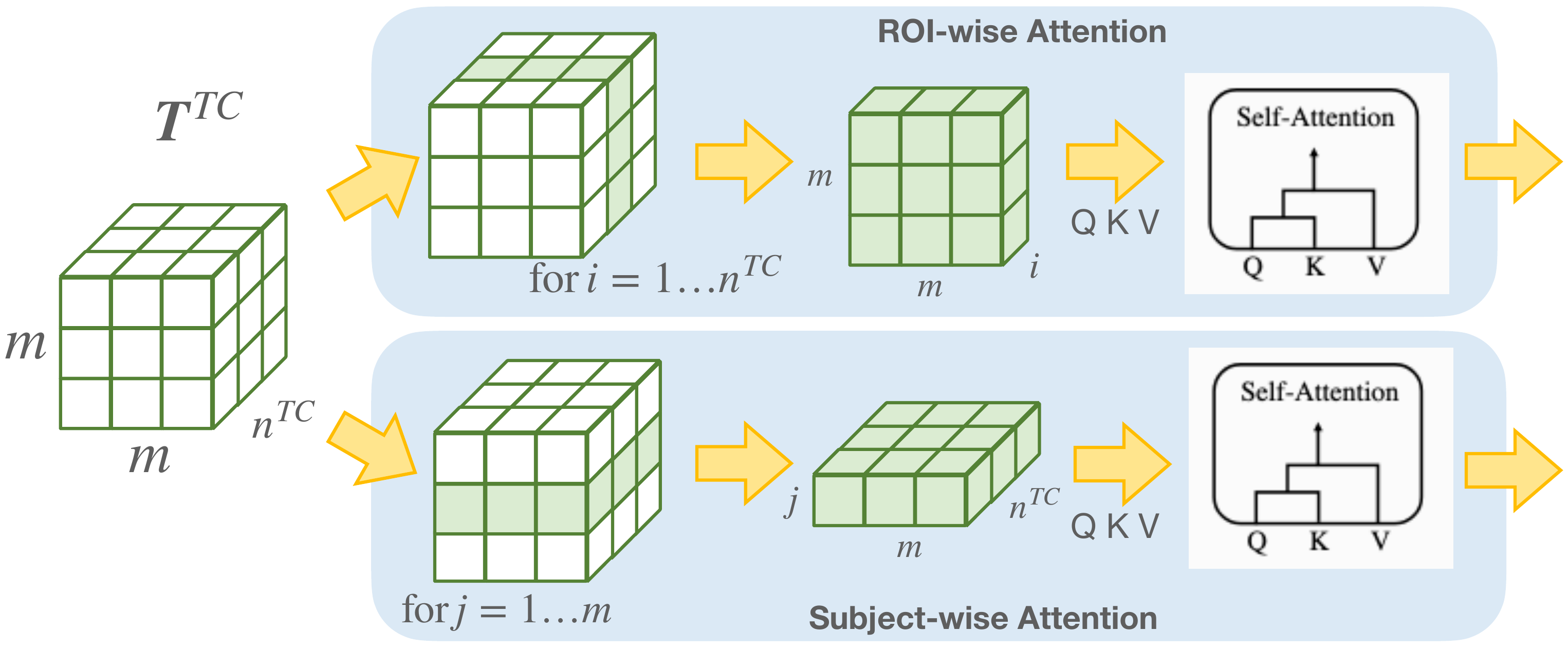}
\end{center}
\caption{The detail of two-stream attention. The ROI- and subject-wise attention blocks compute self-attention from different views of the input. Parameters of self-attention inside these two branches are independent.}
\label{fig:attn}
\end{figure}

\begin{figure*}[h]
\begin{center}
\includegraphics[width=.72\linewidth]{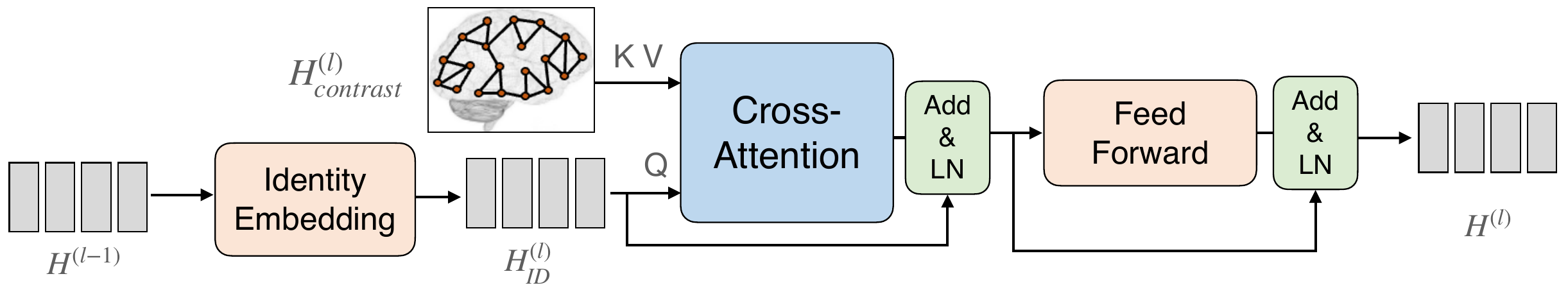}
\end{center}
\caption{The architecture of the cross decoder. The generated contrast graph is incorporated with the identity-embedded brain network by a cross-attention for the downstream representation learning.}
\label{fig:decoder}
\end{figure*}

Taking Autism as an example, we use tensors $\boldsymbol{T}^{TC} \in \mathbb{R}^{n^{TC} \times m \times m}$ and  $\boldsymbol{T}^{ASD} \in \mathbb{R}^{n^{ASD} \times m \times m}$ to denote all brain networks of TC and ASD groups in the training set, respectively.  $n^{TC}$ and $n^{ASD}$ denote the number of subjects in TC/ASD groups and $m$ denotes the number of ROIs. 
As illustrated in Fig. \ref{fig:encoder}, given the input set of brain networks that belong to different groups, the objective of the two-stream attention is to generate a summary graph for each group. An ROI-wise attention and a subject-wise attention are first computed independently for each group of brain networks.
We adopt the self-attention mechanism \cite{vaswani2017attention} for the ROI- and subject-wise attention. 
In general, given a matrix $\boldsymbol{X} \in \mathbb{R}^{k \times d}$, where $k$ and $d$ are arbitrary integers, the self-attention function can be written as:
\begin{equation}
\label{eq:attn}
\small
\operatorname{Attn}(\boldsymbol{X}) = \operatorname{norm} \left(\boldsymbol{X} + \phi\left(\frac{\boldsymbol{Q} \boldsymbol{K}^\mathsf{T}}{\sqrt{k}}\right) \boldsymbol{V}\right),
\end{equation}
\begin{equation}
\label{eq:qkv}
\small
\boldsymbol{Q}=\boldsymbol{X}\boldsymbol{W}_Q, \boldsymbol{K}=\boldsymbol{X}\boldsymbol{W}_K, \boldsymbol{V}=\boldsymbol{X}\boldsymbol{W}_V,
\end{equation}
\begin{equation}
\label{eq:softmax}
\begin{split}
\small
\phi(\boldsymbol{z})_i=\operatorname{softmax}(\boldsymbol{z})_i=\frac{e^{\boldsymbol{z}_i}}{\sum_{j=1}^d e^{\boldsymbol{z}_j}}, \operatorname{for} i = 1 \ldots k, \boldsymbol{z} \in \mathbb{R}^{k},  
\end{split}
\end{equation}
\noindent
where $\boldsymbol{W}_Q, \boldsymbol{W}_K, \boldsymbol{W}_V \in \mathbb{R}^{d \times d}$ are parameter matrices, 
$e$ is the Euler's number, and $\operatorname{norm}(\cdot)$ is a normalization function.

For the brain networks of TC group $\boldsymbol{T}^{TC}$, the detail of two-stream attention is shown in Fig. \ref{fig:attn}.
In ROI-wise attention $\operatorname{Attn}_{ROI}(\cdot)$, for each subject $i \in \{1 \ldots n^{TC}\}$, we compute the self-attention across all ROIs in this subject. The most informative ROIs inside each subject will be extracted in this way. Similarly, in subject-wise attention $\operatorname{Attn}_{subject}(\cdot)$, for each ROI $j = 1 \ldots m$, we compute the self-attention across all subjects in this group. The input of subject-wise attention is implemented by simply transposing the subject and ROI dimensions. This operation helps our model highlight the most discriminative subjects of a certain ROI. The rationale of such attention mechanism is that we intend to extract the group-invariant feature and filter out the SPS noise. 

Note that the normalization function we are using here (in Fig. \ref{fig:encoder}) is batch normalization (BN) instead of the commonly used layer normalization (LN) in Transformer \cite{vaswani2017attention}. It is because (1) the input of the two-stream attention is constant for each training step; (2) the lengths of input sequences are consistent ($m$ for ROI-wise attention, $n^{TC}$ for subject-wise attention); (3) we aim to highlight the consistency within each group. We conduct an ablation study in Section \ref{subsec:ablation} to verify the effectiveness of our model architecture.

Afterward, the summary graph of the TC group $\boldsymbol{H}^{TC}_{sum} \in \mathbb{R}^{m \times m}$ is generated as:
\begin{equation}
\label{eq:summary_graph}
\boldsymbol{H}_{sum}^{TC} = \frac{1}{n^{TC}} \sum^{n^{TC}}_{i=1} (\boldsymbol{H}_{ROI}^{TC} + \boldsymbol{H}_{subject}^{TC})(i,:,:),
\end{equation}
\begin{equation}
\label{eq:attn_roi}
\small
\boldsymbol{H}_{ROI}^{TC} = \left[ \operatorname{Attn}_{ROI} \left( \boldsymbol{T}^{TC}(i,:,:) \right) : i = 1, \dots, n^{TC} \right],  
\end{equation}
\begin{equation}
\label{eq:attn_subject}
\small
\boldsymbol{H}_{subject}^{TC} = \left[ \operatorname{Attn}_{subject} \left( \boldsymbol{T}^{TC}(:,j,:) \right) : j = 1, \dots, m \right], 
\end{equation}
\noindent
where $\boldsymbol{H}_{ROI}^{TC}, \boldsymbol{H}_{subject}^{TC} \in \mathbb{R}^{n^{TC} \times m \times m}$ denote two 3-dimensional matrices to store the outputs of the ROI-wise and subject-wise attention functions, and $\left[ \cdot \right]$ denotes the stack function to combine a set of matrices to a single higher-dimensional matrix. The summary graphs of other groups can also be obtained in a similar way. Once we obtain the summary graphs $\mathcal{H}_{sum} = \{\boldsymbol{H}_{sum}^{TC}, \boldsymbol{H}_{sum}^{ASD}\}$ for TC and ASD groups, the contrast graph is generated by computing the variance of all summary graphs:
\begin{equation}
\label{eq:contrast_graph}
\small
\boldsymbol{H}_{contrast} = \frac{1}{|\mathcal{H}_{sum}|} \sum_{\boldsymbol{H}_{sum} \in \mathcal{H}_{sum}} \left( \boldsymbol{H}_{sum} - \boldsymbol{\bar{H}}_{sum} \right)^2, 
\end{equation}
\begin{equation}
\label{eq:mean}
\small
\boldsymbol{\bar{H}}_{sum} = \frac{1}{|\mathcal{H}_{sum}|} \sum_{\boldsymbol{H}_{sum} \in \mathcal{H}_{sum}} \boldsymbol{H}_{sum}.
\end{equation}
The generated contrast graph $\boldsymbol{H}_{contrast}$ contains the discriminative information about the specific disease, which can be incorporated with the cross decoder to boost the downstream brain network representation learning. The contrast graph generation also works for datasets with multiple groups by making contrast among all groups. Note that we only use subjects in the training set to train the contrast graph encoder. The generated contrast graph is used in the test stage as prior knowledge to avoid data leakage.

\subsection{Cross Decoder with Identity Embedding}
\label{subsec:contrasformer}

By leveraging the group-discriminative information captured in the contrast graph, the cross decoder of Contrasformer aims to produce a high-quality representation for each input brain network (Fig. \ref{fig:decoder}). 

In graph transformer models, positional embedding is commonly used to encode the topological information of the graph. However, existing designs for general graph representation learning, such as distance-based, centrality-based and eigenvector-based positional embedding \cite{li2020distance,ying2021transformers,wang2022equivariant}, can hardly migrate to brain network due to its high density (always fully connected). The correlation-based brain networks naturally contain sufficient positional information for ROIs. Therefore the general-purposed positional embedding is not only expensive but also redundant in our case.

Instead of positional embedding that captures the topological information of the graph structure, we propose a learnable identity embedding to adaptively learn the unique identity for each ROI. Such embedding attaches the same identity for nodes that belong to the same ROI. As shown in Eq. (\ref{eq:id_enc}), we introduce a parameter matrix $\boldsymbol{W}_{ID}^{(l)} \in \mathbb{R}^{m \times m}$ to encode the identity of nodes. $\delta(\cdot)$ denotes a multilayer perceptron (MLP).
\begin{equation}
\label{eq:id_enc}
\small
\boldsymbol{H}_{ID}^{(l)} = \boldsymbol{H}^{(l-1)} + \delta(\boldsymbol{H}^{(l-1)} + \boldsymbol{W}_{ID}^{(l-1)}).
\end{equation}

After identity embedding, we combine the contrast graph with the encoded brain network by a cross-attention function followed by a layer normalization. The encoded brain network $\boldsymbol{H}_{ID}^{(l)}$ serves as $\boldsymbol{Q}$ while the contrast graph $\boldsymbol{H}_{contrast}^{(l)}$ is treated as $\boldsymbol{K}$ and $\boldsymbol{V}$ in Eq. (\ref{eq:attn}). Each input brain network is fed into the cross-attention module to query the task-specific information in the contrast graph. The intuition here is to use the contrast graph as prior knowledge to guide the brain network representation learning. \jiaxing{By hiding the non-indicative ROIs/connections and emphasizing the indicative ones, task-specific domain knowledge is introduced to the embedded brain networks.}

In addition to the cross-attention sub-layer, a position-wise feed-forward network (FFN) with a layer normalization is applied to each position to get the output node representations $\boldsymbol{H}^{(l)}$ of the $l$-th Contrasformer layer. The FFN is applied to each position separately and identically \cite{vaswani2017attention}.
After $L$ layers of Contrasformer, a readout function $ \operatorname{Readout}(\cdot)$ is applied to the node representations to generate a graph representation: $\boldsymbol{hg} = \operatorname{Readout}(\boldsymbol{H}^{(l)})$. The graph representation is then passed to the prediction head for classification.


\subsection{Loss Functions}
\label{subsec:loss}

In order to introduce domain knowledge and make model optimization easier to converge, we utilize 4 loss functions to guide the end-to-end training. (1) A commonly-used cross-entropy loss $\mathcal{L}_{cls}$ \cite{cox1958regression} for graph classification; (2) an entropy loss $\mathcal{L}_{entropy}$ for contrast graph sparsification; (3) a cluster loss $\mathcal{L}_{cluster}$ to take the group relationship (i.e., similarity and discrepancy) into account; (4) a contrastive loss $\mathcal{L}_{contrast}$ to constrain node-identity awareness. The total loss is computed by:
\begin{equation}
\label{eq:total_loss}
\small
\mathcal{L}_{total} = \mathcal{L}_{cls} + \lambda_1 * \mathcal{L}_{entropy} + \lambda_2 * \mathcal{L}_{cluster} + \lambda_3 * \mathcal{L}_{contrast},
\end{equation} 
\noindent
where $\lambda_1$, $\lambda_2$ and $\lambda_3$ are trainable trade-off hyperparameters.

\begin{figure}[ht]
\begin{center}
\includegraphics[width=\linewidth]{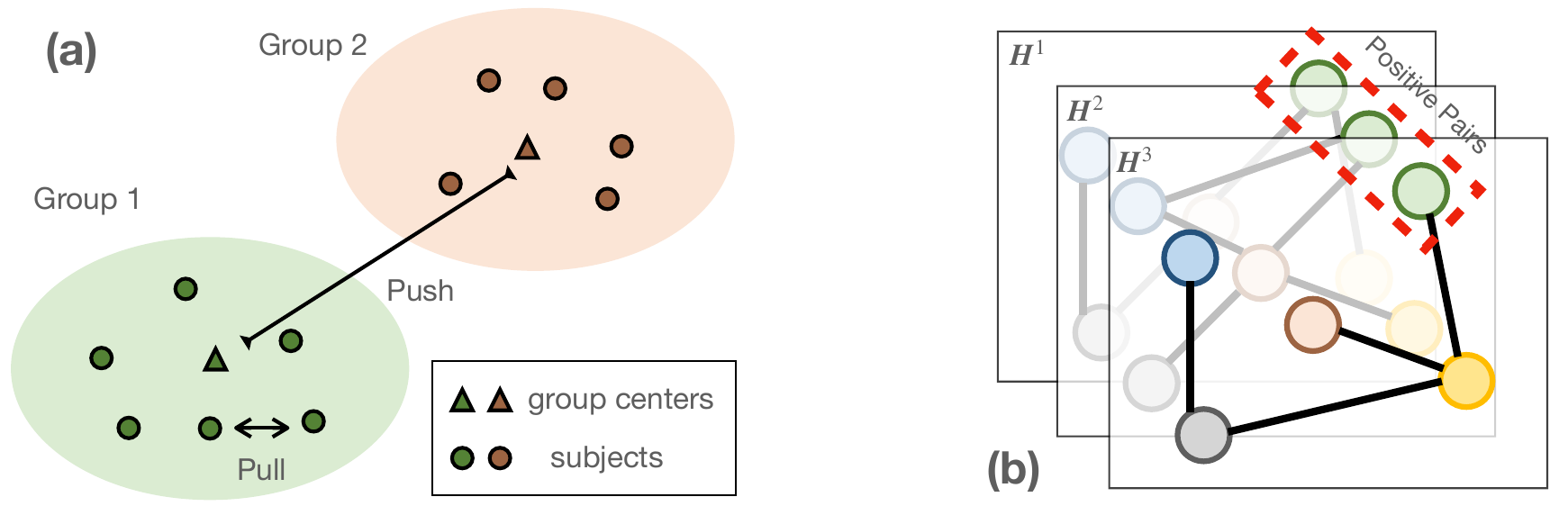}
\end{center}
\caption{(a) The cluster loss enforces subjects that belong to the same group to get similar representations, the subjects from different groups less similar. (b) The contrastive loss treats the nodes belonging to the same ROI as positive pairs, and all the other node pairs are considered negative pairs.}
\label{fig:losses}
\end{figure}

\noindent
\textbf{Entropy Loss.} To prevent a smooth contrast graph that treats all ROIs equally, risking the loss of discriminative ability, we introduce a sparsity constraint. To achieve this, we employ an entropy loss, compelling the model to prioritize the most task-specific ROI connections. The entropy loss is formulated as follows:
\begin{equation}
\label{eq:entropy_loss}
\small
\mathcal{L}_{entropy} = \frac{1}{m} \sum^m_{i=1} \operatorname{entropy} \left( \boldsymbol{H}_{contrast}(i,:) \right),
\end{equation}
\begin{equation}
\label{eq:entropy}
\small
\operatorname{entropy}(\boldsymbol{p}) = - \sum_{j=1}^m \boldsymbol{p}_j \log(\boldsymbol{p}_j).
\end{equation}
\noindent
\textbf{Cluster Loss.} 
Most existing GNN/Transformer architectures treat individual input graphs independently during training. Neglecting the relationships between classes could lead to a significant compromise in model effectiveness for downstream classification tasks \cite{XU2024121061}. For our application, we want to find the common patterns/biomarkers for a certain neurological disorder identification task.
Thus we propose a cluster loss to leverage graph-level similarity and make the graph representations more sepratable:
\begin{equation}
\label{eq:cluster_loss}
\small
\mathcal{L}_{cluster} =  \operatorname{log} \frac {\operatorname{exp}(\sum_{c \in \mathcal{C}} \sigma_c^2)} {\operatorname{exp}(\sum_{c \in \mathcal{C}} \sum_{i \in \mathcal{C}} ||\mu_c -\mu_i||_2)} ,
\end{equation}
\begin{equation}
\label{eq:mu_sigma}
\small
\mu_c = \sum_{k \in \mathcal{S}^c} \frac{\boldsymbol{hg}^k}{|\mathcal{S}^c|}, \ \ \ \sigma_c^2 = \sum_{k \in \mathcal{S}^c} \frac{(\boldsymbol{hg}^k - \mu_c)^2}{|\mathcal{S}^c|},
\end{equation}
\noindent
where $\mu_c$ and $\sigma_c$ denote the mean and standard deviation of graph representations belonging to group $c$, $\mathcal{S}^c$ denotes the subject indices that belong to group $c$, $\mathcal{C}$ denotes the set of classes, and $\boldsymbol{hg}^k$ denotes the graph representation with index $k$ in $\mathcal{S}^c$.
As in Fig. \ref{fig:losses}(a), the cluster loss aims to pull the graph representations within a group close to each other and push the centers of groups as far as possible. By using such cluster loss, the group-level relationship of all classes is considered equally no matter how many subjects it contains. 

\noindent
\textbf{Contrastive Loss.} Existing graph contrastive learning technologies \cite{you2020graph, zhu2021graph} require data argumentation by modifying graph structure or dropping node/edge features. Such contrast is still limited to each individual graph. It also cannot be migrated to our brain networks because the connectivity matrix naturally contains its structural and positional information.

Herein, we design an ROI-level contrastive loss to further leverage the node identity of ROIs. Thanks to the unique property of brain networks' node-identity awareness, we are able to conduct contrastive learning by aligning ROIs across subjects. \jiaxing{To the best of our knowledge, this is the first attempt to bring in contrastive constraints at the ROI level for brain network analysis.} The contrastive loss $\mathcal{L}_{contrast}$ shown in Eq. (\ref{eq:contrast_loss}) is defined to enforce maximizing the consistency between positive pairs compared with negative pairs \cite{chen2020simple}. 
We use $\boldsymbol{H}^i_j$ to denote the node representation for the $j$-th node in the $i$-th subject, where $j = 1, \dots, m$, $i =  1, \dots, n$, and $n$ is the total number of subjects in the training set. We denote the set of positive pairs as $\mathcal{P}^{pos} = \{ (\boldsymbol{H}^i_j, \boldsymbol{H}^p_j) \}$ and the set of negative pairs as $\mathcal{P}^{neg} = \{ (\boldsymbol{H}^i_j, \boldsymbol{H}^q_r) \}$, where $p =  1, \dots, n, p \neq i$, $q =  1, \dots, n$, $r =  1, \dots, m, r \neq j$. A temperature hyper-parameter $\tau$ \cite{you2020graph} is introduced to control the smoothness of the probability distribution, and $\operatorname{sim}(\cdot)$ denotes the cosine similarity function.
As elaborated in Fig. \ref{fig:losses}(b), we treat the same ROI of all subjects as positive pairs, and different ROIs from the same/different subjects as negative pairs to emphasize the node-identity awareness of brain networks. 
\begin{equation}
\label{eq:contrast_loss}
\small
\mathcal{L}_{contrast} = - \operatorname{log} \frac {\sum_{(\boldsymbol{H}^i_j, \boldsymbol{H}^p_j) \in \mathcal{P}^{pos}} \operatorname{exp}(\operatorname{sim}(\boldsymbol{H}^i_j, \boldsymbol{H}^p_j) / \tau)} {\sum_{(\boldsymbol{H}^i_j, \boldsymbol{H}^q_r) \in \mathcal{P}^{neg}} \operatorname{exp} (\operatorname{sim}(\boldsymbol{H}^i_j, \boldsymbol{H}^q_r) / \tau)}.
\end{equation}

\section{Experimental Study}
\label{sec:experiments}

\begin{table}[h]
\centering
\small
\caption{Statistics of Brain Network Datasets.}
\begin{tabular}{cccccc}
\hline
Dataset & Condition & Site\# & Graph\# & Avg BOLD length & Class\# \\ \hline
Mātai            & mTBI    & 1      & 60  & 198 & 2  \\
PPMI             & PD    & 21      & 209  & 198 & 4  \\
ADNI             & AD     & 89    & 1326 & 344 & 6  \\
ABIDE            & ASD      & 17   & 1025 & 201 & 2 \\ \hline
\end{tabular}
\label{tab:dataset_statistic}
\end{table}

\begin{table*}[h]
\caption{Graph Classification Results (Average Accuracy ± Standard Deviation) over 10-fold-CV. The best result is highlighted in \textbf{bold}. The second-best result is \underline{underlined}.}
\label{tab:main_results}
\centering
\small
\begin{tabular}{cccccc}
\hline
          & Model   & Mātai         & PPMI          & ADNI         & ABIDE        \\ \hline
\multirow{2}{*}{\shortstack{\textit{Conventional}\\ \textit{ML methods}}} &  LR            & 60.00 ± 20.00 & 56.48 ± 6.76  & 61.97 ± 4.24 & 64.81 ± 3.70 \\
& SVM           & 56.67 ± 17.00 & 63.21 ± 8.62  & 61.52 ± 4.95 & 64.41 ± 5.09 \\ \hline
\multirow{5}{*}{\shortstack{\textit{General-}\\ \textit{purposed}\\ \textit{GNNs}}} & GCN           & 56.67 ± 18.56 & 54.02 ± 9.06  & 60.40 ± 4.89 & 60.19 ± 2.96 \\
& GraphSAGE           & 61.67 ± 10.67 & 55.00 ± 12.89  & 59.35 ± 3.39 & 61.75 ± 4.35 \\
& GAT           & 66.67 ± 18.33 & 54.98 ± 8.03  & 59.73 ± 2.85 & 60.10 ± 4.13 \\
& GatedGCN      & 58.33 ± 8.33  & 52.60 ± 11.51 & 64.55 ± 1.87 & 61.66 ± 3.36 \\
& GPS           & \underline{63.33} ± 14.53 & 57.50 ± 7.83  & 65.17 ± 3.50 & 63.04 ± 3.36 \\ \hline
\multirow{6}{*}{\shortstack{\textit{Neural networks}\\ \textit{tailored for}\\ \textit{brain networks}} } &  BrainNetCNN   & 61.67 ± 13.33 & 57.33 ± 10.32 & 60.48 ± 3.29 & \underline{65.75} ± 3.24 \\
& LiNet         & 51.67 ± 13.84 & 60.71 ± 10.61 & 61.91 ± 1.02 & 54.05 ± 4.50 \\
& PRGNN         & 55.00 ± 16.75 & 58.83 ± 6.89  & 62.51 ± 3.36 & 59.71 ± 4.54 \\
& BrainGNN      & 53.33 ± 24.49 & 61.71 ± 6.05  & 61.05 ± 1.23 & 62.88 ± 2.46 \\
& BNTF          & 61.67 ± 11.17 & 51.60 ± 6.15  & 65.49 ± 3.25 & 63.70 ± 4.84 \\
& ContrastPool  & 61.67 ± 13.02 & \underline{64.00} ± 6.63  & \underline{65.67} ± 6.64 & 65.01 ± 3.84 \\ \hline
\textit{Ours} &  Contrasformer & \textbf{68.33} ± 18.93 & \textbf{67.00} ± 4.58  & \textbf{69.33} ± 3.63 & \textbf{66.27} ± 3.67 \\ \hline
\end{tabular}
\end{table*}

\begin{table*}[h]
\caption{Results of more evaluation metrics on ABIDE. The best result is highlighted in \textbf{bold}. The second-best result is \underline{underlined}.}
\label{tab:other_metrics}
\centering
\small
\begin{tabular}{cccccc}
\hline
 & Model        & Precision    & Recall        & micro-F1     & ROC-AUC      \\ \hline
\multirow{5}{*}{\shortstack{\textit{General-}\\ \textit{purposed}\\ \textit{GNNs}}} &
  GCN & 59.69 ± 5.50 & 57.56 ± 7.01 & 58.49 ± 5.78 & 61.08 ± 4.92 \\
 & GraphSAGE          & 60.33 ± 4.91 & 58.20 ± 7.85  & 58.99 ± 5.45 & 61.59 ± 4.36 \\
 & GAT          & 59.57 ± 4.63 & 55.11 ± 7.89  & 56.96 ± 5.16 & 60.43 ± 3.88 \\
 & GatedGCN     & 61.65 ± 4.12 & 55.74 ± 11.58 & 58.05 ± 8.20 & 62.31 ± 4.32 \\
 & GPS          & 59.97 ± 5.36 & 68.75 ± 11.22 & 63.48 ± 5.98 & 63.34 ± 5.15 \\ \hline
\multirow{6}{*}{\shortstack{\textit{Neural networks}\\ \textit{tailored for}\\ \textit{brain networks}} } &
  BrainNetCNN & \textbf{63.98} ± 3.47 & 61.25 ± 5.66 & 62.39 ± 3.13 & \underline{64.78} ± 2.52 \\
 & LiNet        & 56.34 ± 6.94 & 30.37 ± 9.23  & 38.66 ± 8.32 & 54.39 ± 3.29 \\
 & PRGNN        & 60.83 ± 7.44 & 53.68 ± 8.36  & 56.77 ± 7.15 & 61.01 ± 5.74 \\
 & BrainGNN     & 62.48 ± 5.92 & 57.15 ± 4.66  & 59.42 ± 3.23 & 62.74 ± 3.58 \\
 & BNTF         & 60.34 ± 5.40 & \underline{70.19} ± 8.66  & \underline{64.64} ± 5.65 & 64.15 ± 5.42 \\
 & ContrastPool & \underline{63.56} ± 3.62 & 61.45 ± 5.43  & 62.28 ± 2.81 & 64.52 ± 2.30 \\ \hline
\textit{Ours} &
  Contrasformer &
  62.59 ± 4.40 &
  \textbf{73.07} ± 4.69 &
  \textbf{67.25} ± 3.01 &
  \textbf{66.62} ± 3.47 \\ \hline
\end{tabular}
\end{table*}

\subsection{Brain Network Datasets}

We use four brain network datasets from different data sources for various disorders, which are Mātai for mild traumatic brain injury (mTBI) \cite{xu2023data}, PPMI \cite{badea2017exploring} for Parkinson's disease (PD), ADNI \cite{dadi2019benchmarking} for Alzheimer's disease (AD), and ABIDE \cite{craddock2013neuro} for Autism (ASD). Statistics of the brain network datasets are summarized in Table \ref{tab:dataset_statistic}.

\noindent
\textbf{Mātai}  
Mātai is a longitudinal single site, single scanner study designed for detecting subtle changes in the brain due to a season of playing contact sports. This dataset consists of the brain networks preprocessed from the data collected from Gisborne-Tairāwhiti area, New Zealand, with 35 contact sport players imaged at pre-season (N=35) and post-season (N=25) with subtle brain changes confirmed using diffusion imaging study due to playing contact sports.

\noindent
\textbf{PPMI} The PPMI\footnote{\url{https://www.ppmi- info.org/accessdata-specimens/download-data}} is a comprehensive study aiming to identify biological markers associated with Parkinson's risk, onset, and progression. PPMI comprises multimodal and multi-site MRI images. The dataset consists of subjects from 4 distinct classes: normal control (NC), scans without evidence of dopaminergic deficit (SWEDD), prodromal, and PD. 

\noindent
\textbf{ADNI} The ADNI raw images used in this paper were obtained from the ADNI database (\url{adni.loni.usc.edu}). The ADNI was launched in 2003 as a public-private partnership, led by Principal Investigator Michael W. Weiner, MD. The primary goal of ADNI has been to test whether serial magnetic resonance imaging (MRI), positron emission tomography (PET), other biological markers, and clinical and neuropsychological assessment can be combined to measure the progression of mild cognitive impairment (MCI) and early Alzheimer’s disease (AD). For up-to-date information, see \url{www.adni-info.org}. We include subjects from 6 different stages of AD, from cognitive normal (CN), significant memory concern (SMC), mild cognitive impairment (MCI), early MCI (EMCI), late MCI (LMCI) to AD.

\noindent
\textbf{ABIDE} The ABIDE initiative supports the research on ASD by aggregating functional brain imaging data from laboratories worldwide. ASD is characterized by stereotyped behaviors, including irritability, hyperactivity, and anxiety. Subjects in the dataset are classified into two groups: TC and individuals diagnosed with ASD. 

\subsection{Baseline Models}

We employ diverse baseline models to benchmark our approach, including (1) Conventional machine learning models: Logistic Regression (\textbf{LR}) and Support Vector Machine Classifier (\textbf{SVM}) from scikit-learn \cite{pedregosa2011scikit}. These models take the flattened connectivity matrix as vector input, instead of using the brain network.
(2)General-purposed GNNs: \textbf{GCN} \cite{kipf2016semi}, a mean pooling baseline with a graph convolution network as a message-passing layer; \textbf{GraphSAGE} \cite{hamilton2017inductive} is a mean pooling baseline, which adopts sampling to obtain a fixed number of neighbors for each node; \textbf{GAT} \cite{velivckovic2017graph}, a mean pooling baseline utilizing an attention mechanism to learn relative node-neighbor weights; \textbf{GatedGCN} \cite{bresson2017residual} incorporates gate mechanisms to selectively control information flow during message passing, enabling improved modeling of long-range dependencies in graph; and \textbf{GPS} \cite{rampavsek2022recipe}, a state-of-the-art Transformer-based model incorporating various positional/structural embeddings. 
(3) Neural networks tailored for brain networks: \textbf{BrainNetCNN} \cite{kawahara2017brainnetcnn}, the pioneering CNN regressor for connectome data; \textbf{LiNet} \cite{li2019graph}, an inductive GNN for task-fMRI properties identification; \textbf{PRGNN} \cite{li2020pooling}, a graph pooling method ensuring ROI-selection coherence; \textbf{BrainGNN} \cite{li2021braingnn}, a GNN method with ROI-aware convolutional layers for fMRI information integration; Brain Network Transformer (\textbf{BNTF}) \cite{kan2022brain}, a specialized graph Transformer with orthonormal clustering for brain network analysis; and \textbf{ContrastPool} \cite{10508252}, a node clustering pooling using a dual-attention block for domain-specific information capturing. 

Note that, to prevent over-smoothing in GNNs, we sparsify the adjacency matrices of all input graphs by retaining the top 20\% edges with the highest correlations in $\boldsymbol{M}$.

\begin{table*}[h]
\caption{Results when generalizing to unseen sites. The best result is highlighted in \textbf{bold}. The second-best result is \underline{underlined}.}
\label{tab:unseen_sites}
\small
\centering
\begin{tabular}{l|ccccc}
\hline
              & Accuracy     & Precision    & Recall        & micro-F1     & ROC-AUC      \\ \hline
GCN           & 55.15 ± 3.63 & 48.82 ± 3.54 & 56.67 ± 7.04  & 52.33 ± 4.69 & 55.32 ± 3.76 \\
GatedGCN     & 59.03 ± 1.55  & 52.87 ± 1.67 & 55.33 ± 8.52 & 53.79 ± 4.96 & 58.61 ± 2.17 \\
BrainNetCNN   & \underline{61.55} ± 2.31 & \underline{54.44} ± 2.10 & \underline{74.44} ± 5.19  & \underline{62.81} ± 2.45 & \underline{63.00} ± 2.28 \\
LiNet         & 53.50 ± 2.89 & 45.99 ± 4.07 & 35.11 ± 3.69  & 39.71 ± 3.26 & 51.43 ± 2.70 \\
PRGNN         & 55.63 ± 2.34 & 49.21 ± 2.77 & 49.56 ± 3.86  & 49.35 ± 3.08 & 54.95 ± 2.43 \\
ContrastPool  & 56.18 ± 2.01 & 48.60 ± 3.91 & 42.27 ± 10.77 & 44.61 ± 8.18 & 54.50 ± 2.61 \\
Contrasformer (ours) & \textbf{64.61} ± 1.83 & \textbf{56.52} ± 1.56 & \textbf{77.95} ± 3.53 & \textbf{65.50} ± 1.93 & \textbf{66.22} ± 1.88 \\ \hline
\end{tabular}
\end{table*}

\subsection{Implementation Details}
\label{sec:imple_detail}

In Contrasformer, we adopt a Linear layer for input embedding, a mean pooling layer as the readout function $\operatorname{Readout(\cdot)}$, and a two-layer MLP with ReLU as the prediction head. The temperature hyper-parameter $\tau$ in Eq. (\ref{eq:contrast_loss}) is set to 0.02. 
The settings of our experiments mainly follow those in \cite{dwivedi2020benchmarkgnns}. We split each dataset into 8:1:1 for training, validation and test, respectively. We evaluate each model with the same random seed under 10-fold cross-validation and report the average accuracy. 
The whole network is trained in an end-to-end manner using the Adam optimizer \cite{kingma2014adam}. We use the early stopping criterion, i.e., we stop the training once there is no further improvement on the validation loss during 25 epochs. 
All experiments were conducted on a Linux server with an Intel(R) Core(TM) i9-10940X CPU (3.30GHz), a GeForce GTX 3090 GPU, and a 125GB RAM.

\subsection{Main Results}

We report the accuracy on 4 brain network datasets in Table \ref{tab:main_results}. Our proposed Contrasformer consistently outperforms all 13 baselines on all datasets. 
In particular, Contrasformer improves over all networks specifically designed for brain networks on these four datasets by up to 10.8\% ((68.33\%-61.67\%)/61.67\% = 10.8\% on Mātai).
These experimental results demonstrate the effectiveness of our brain network oriented model design. 
The improvement may result from two reasons. First, the participation of the contrast graph in brain network representation learning provides reasonable and discriminative information about certain conditions. Second, the properties of fMRI and group constraints are introduced to the model training by dedicated loss functions.

Apart from accuracy, we also report other evaluation metrics, including precision, recall, micro-F1, and ROC-AUC, of all the models on the ABIDE dataset. As shown in Table \ref{tab:other_metrics}, Contrasformer performs the best over all these metrics except for precision. 
We can also discover that compared with other baselines, our Contrasformer can dramatically improve recall without sacrificing precision. Besides, in medical diagnostics, it's crucial to ensure that all individuals with a certain condition are correctly identified, even if it means some false positives. Missing a true positive (failing to diagnose a disease) can have severe consequences, while false positives can be further examined or retested. Therefore, models with higher recall rates, like our Contrasformer, are more suitable for application in real-life medical auxiliary diagnosis. 

\subsection{Model Interpretation}

\begin{figure}[h]
  \centering
\includegraphics[width=\linewidth]{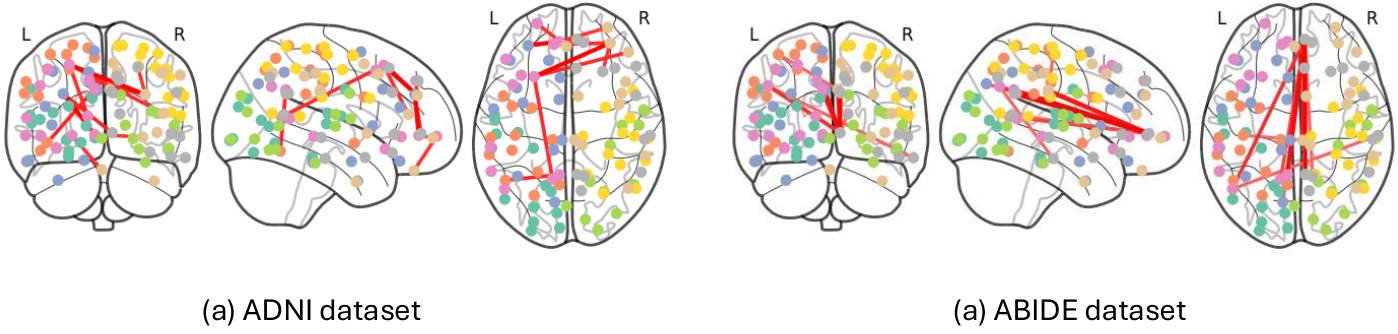}
  \caption{Contrast graph visualization by highlighting the top 10 edges with the highest strength.}
  \label{fig:contrast_graph}
\end{figure}

\noindent
\textbf{Contrast Graph Visualization.} \jiaxing{Despite the high accuracy achieved by our model, a critical concern is the interpretability of their decision-making process. In the context of brain biomarker detection, identifying salient ROIs associated with predictions as candidate/potential biomarkers is crucial. In this study, we leverage built-in model interpretability to explore disease-specific biomarker analysis.}
To interpret Contrasformer's reasoning, we visualize the learnt contrast graphs for Alzheimer’s Disease and Autism on the ADNI and ABIDE datasets, by using the Nilearn toolbox \cite{abraham2014machine}. We select edges with the top 10 attention weights. As depicted in Fig. \ref{fig:contrast_graph}(a), highlighted connections between the lateral prefrontal cortex, prefrontal cortex, and dorsal prefrontal cortex medial prefrontal cortex in the ADNI dataset suggest potential AD-specific neural mechanisms \cite{li2012attention}. These regions have been recognized as key areas in previous AD studies \cite{rombouts2002alterations,frisch2013dissociating}. Similar ROI-wise interpretations are found in Autism. In Fig. \ref{fig:contrast_graph}(b), highlighted ROIs related to precuneus posterior cingulate cortex, cingulate, and dorsal prefrontal cortex medial prefrontal cortex on ABIDE align with domain knowledge from prior Autism research \cite{assaf2010abnormal,ecker2010describing,kana2017neural}.

\begin{table*}[h]
\caption{Ablation study on important modules/loss functions in Contrasformer on ABIDE dataset.}
\label{tab:ablation_component}
\centering
\small
\begin{tabular}{cccc|ccccc}
\hline
$\operatorname{Attn}_{ROI}$ & $\operatorname{Attn}_{subject}$ & batch norm & ID enc & Accuracy & Precision & Recall & micro-F1 & ROC-AUC \\ \hline
  & \checkmark & \checkmark & \checkmark & 63.14 ± 4.71 & 59.18 ± 5.36 & 75.76 ± 7.83  & 66.04 ± 4.11 & 63.77 ± 4.78 \\
\checkmark &   & \checkmark & \checkmark & 63.63 ± 4.28 & 62.42 ± 6.02 & 60.00 ± 6.37  & 60.91 ± 4.79 & 63.44 ± 4.03 \\
\checkmark & \checkmark &   & \checkmark & 65.69 ± 3.90 & \textbf{62.83} ± 4.36 & 68.90 ± 10.24 & 65.22 ± 5.47 & 65.83 ± 3.91 \\
\checkmark & \checkmark & \checkmark &   & 64.22 ± 3.87 & 59.97 ± 5.41 & \textbf{76.80} ± 4.51  & 67.12 ± 3.31 & 64.84 ± 4.60 \\
\checkmark & \checkmark & \checkmark & \checkmark & \textbf{66.27} ± 3.67 & 62.59 ± 4.40 & 73.07 ± 4.69  & \textbf{67.25} ± 3.01 & \textbf{66.62} ± 3.47 \\ \hline
\hline
$\mathcal{L}_{cls}$ & $\mathcal{L}_{entropy}$ & $\mathcal{L}_{cluster}$ & $\mathcal{L}_{contrast}$ & Accuracy     & Precision    & Recall        & micro-F1     & ROC-AUC      \\ \hline
\checkmark      &            & \checkmark          & \checkmark           & 64.61 ± 3.93 & 61.79 ± 4.63 & 68.51 ± 9.36  & 64.45 ± 4.80 & 64.80 ± 3.69 \\
\checkmark      & \checkmark          &            & \checkmark           & 64.41 ± 4.34 & 60.36 ± 4.90 & \textbf{75.57} ± 5.44  & 66.84 ± 2.86 & 64.99 ± 4.00 \\
\checkmark      & \checkmark          & \checkmark          &             & 59.51 ± 4.09 & 57.66 ± 6.84 & 58.25 ± 13.77 & 56.98 ± 8.30 & 59.50 ± 5.72 \\
\checkmark      & \checkmark          & \checkmark          & \checkmark           & \textbf{66.27} ± 3.67 & \textbf{62.59} ± 4.40 & 73.07 ± 4.69  & \textbf{67.25} ± 3.01 & \textbf{66.62} ± 3.47 \\ \hline
\end{tabular}
\end{table*}

\noindent
\textbf{Generalization Ability.} \jiaxing{While task-specific biomarkers are valuable for identifying disease-relevant features, it is crucial to determine whether these biomarkers are invariant over the entire population, i.e., whether they generalize well across sub-populations and other diverse populations.}
To assess the generalization ability of Contrasformer, we conduct evaluations on subjects from previously unseen sites. Specifically, we designate two sites from the ABIDE dataset as the test set, while the remaining subjects are split into training and validation sets, maintaining an 8:1:1 ratio. The test set remains constant across 10 experiments with different train-validation splits, and the average results are reported in Table \ref{tab:unseen_sites}. Notably, Contrasformer consistently outperforms all baseline models, demonstrating its robustness against SPS noise. The baseline models exhibit a significant performance reduction compared to Table \ref{tab:main_results}, indicating the detrimental impact of SPS noise on their generalization ability. In contrast, Contrasformer, with its two-stream attention mechanism, effectively extracts and emphasizes task-related features, mitigating the adverse effects of SPS noise.

\subsection{Ablation Study}
\label{subsec:ablation}

In this subsection, we empirically validate the design of (1) the important modules; and (2) the loss functions. All experiments in this subsection are conducted on ABIDE dataset.

\noindent
\textbf{Important Modules.} To inspect the effect of the important modules, we conduct experiments by disabling each of them without modifying other settings. The results are reported in Table \ref{tab:ablation_component}. For ROI-wise attention $\operatorname{Attn}_{ROI}$, subject-wise attention $\operatorname{Attn}_{subject}$, and identity encoding (denoted as ``ID enc'' in the table), we disable them by simply removing these modules. When disabling ``batch norm'', we replace the batch normalization functions in the two-stream attention by layer normalizations. The results demonstrate that Contrasformer with all important modules enabled achieves the best performance. Besides, the experiment of disabling $\operatorname{Attn}_{subject}$ indicates that the outstanding recall of Contrasformer is mainly contributed by the subject-wise attention, demonstrating the effectiveness of extracting discriminative ROIs across subjects.

\noindent
\textbf{Loss Functions.} To verify the effectiveness of our proposed losses, we test our design of the loss functions by disabling them one by one. As shown in Table \ref{tab:ablation_component}, the results demonstrate that all of those three auxiliary losses are effective in boosting the model performance. Besides, we find that the most important one is the contrastive loss. This observation indicates the necessity of introducing the constraint of node awareness.

\section{Conclusion and Future Works}
\label{sec:conclusion}

To overcome the hurdles of SPS noise and node-identity awareness, we introduce Contrasformer, a contrastive brain network Transformer. Through a contrast graph encoder with two-stream attention and a cross decoder with identity embedding, Contrasformer adaptly handles SPS noise, enhances node identity awareness, and captures group-specific patterns. Our model outperforms state-of-the-art methods in identifying neurological disorders across diverse datasets. The improvement over all the best models specifically designed for brain networks is up to 10.8\%. Beyond superior performance, Contrasformer provides interpretable insights aligned with neuroscience literature. This work marks a significant advancement in harnessing Transformer models for fMRI-based brain network analysis, opening avenues for deeper understanding and diagnosis of neurological conditions. In the future, we plan to extend our model to various modalities of medical imaging, such as Diffusion Tensor Imaging (DTI), and explore the multi-modal solution for neurological disorder identification.

\section{Acknowledgments}

This research/project is supported by the National Research Foundation, Singapore under its Industry Alignment Fund – Pre-positioning (IAF-PP) Funding Initiative, and the Ministry of Education, Singapore under its MOE Academic Research Fund Tier 2 (STEM RIE2025 Award MOE-T2EP20220-0006), and MBIE Catalyst: Strategic Fund NZ-Singapore Data Science Research Programme UOAX2001. Any opinions, findings and conclusions or recommendations expressed in this material are those of the author(s) and do not reflect the views of National Research Foundation, Singapore, and the Ministry of Education, Singapore.

Part of the data used in preparation of this article were obtained from the Alzheimer’s Disease Neuroimaging Initiative (ADNI) database (adni.loni.usc.edu). As such, the investigators within the ADNI contributed to the design and implementation of ADNI and/or provided data but did not participate in the analysis or writing of this report. A complete listing of ADNI investigators can be found at: \url{http://adni.loni.usc.edu/wp-content/uploads/how_to_apply/ADNI_Acknowledgement_List.pdf}.

\bibliographystyle{ACM-Reference-Format}
\bibliography{ref}

\end{document}